\pdfoutput=1

\documentclass[11pt]{article}

\usepackage[]{EMNLP2023}

\usepackage{times}
\usepackage{amsmath}
\usepackage{bbm}
\usepackage{amssymb}
\usepackage{dsfont}
\usepackage[utf8]{inputenc} 
\usepackage[T1]{fontenc}    
\usepackage{hyperref}       
\usepackage{url}            
\usepackage{booktabs}       
\usepackage{amsfonts}       
\usepackage{nicefrac}       
\usepackage{microtype}      
 \usepackage{multirow}
 \usepackage{graphicx}
\usepackage{latexsym}
\usepackage{adjustbox}

\usepackage{amsmath}
\usepackage{graphicx}
\usepackage[T1]{fontenc}

\usepackage[utf8]{inputenc}

\usepackage{microtype}

\usepackage{inconsolata}
\usepackage{fancyhdr}

\title{Navigating Uncertainty: Optimizing API Dependency for Hallucination Reduction in Closed-Book Question Answering}

\author{Pierre Erbacher\thanks{* Authors contributed equally to this work} \\
  Sorbonne Université\\
  \texttt{pierre.erbacher@isir.upmc.fr} \\\And
  Louis Falissar\footnotemark[1] \\
  Sorbonne Université \\
  \texttt{louis.falissard@isir.upmc.fr}  \\\AND
  Vincent Guigue \\
  AgroParisTech\\
  \texttt{vincent.guigue@agroparistech.fr}  \\\And
  Laure Soulier \\
  Sorbonne Université\\
  \texttt{laure.soulier@isir.upmc.fr} \\}

\begin{document}
\maketitle
\thispagestyle{fancy}
\begin{abstract}
While Large Language Models (LLM) are able to accumulate and restore knowledge, they are still prone to hallucination. Especially when faced with factual questions, LLM cannot only rely on knowledge stored in parameters to guarantee truthful and correct answers. Augmenting these models with the ability to search on external information sources, such as the web, is a promising approach to ground knowledge to retrieve information. However, searching in a large collection of documents introduces additional computational/time costs. An optimal behavior would be to query external resources only when the LLM is not confident about answers. In this paper, we propose a new LLM able to self-estimate if it is able to answer directly or needs to request an external tool. We investigate a supervised approach by introducing a hallucination masking mechanism in which labels are generated using a close book question-answering task. In addition, we propose to leverage parameter-efficient fine-tuning techniques to train our model on a small amount of data. Our model directly provides answers for $78.2\%$ of the known queries and opts to search for $77.2\%$ of the unknown ones. This results in the API being utilized only $62\%$ of the time.
\end{abstract}

\section{Introduction}
\begin{figure}[ht]
    \centering
    \includegraphics[scale=0.43]{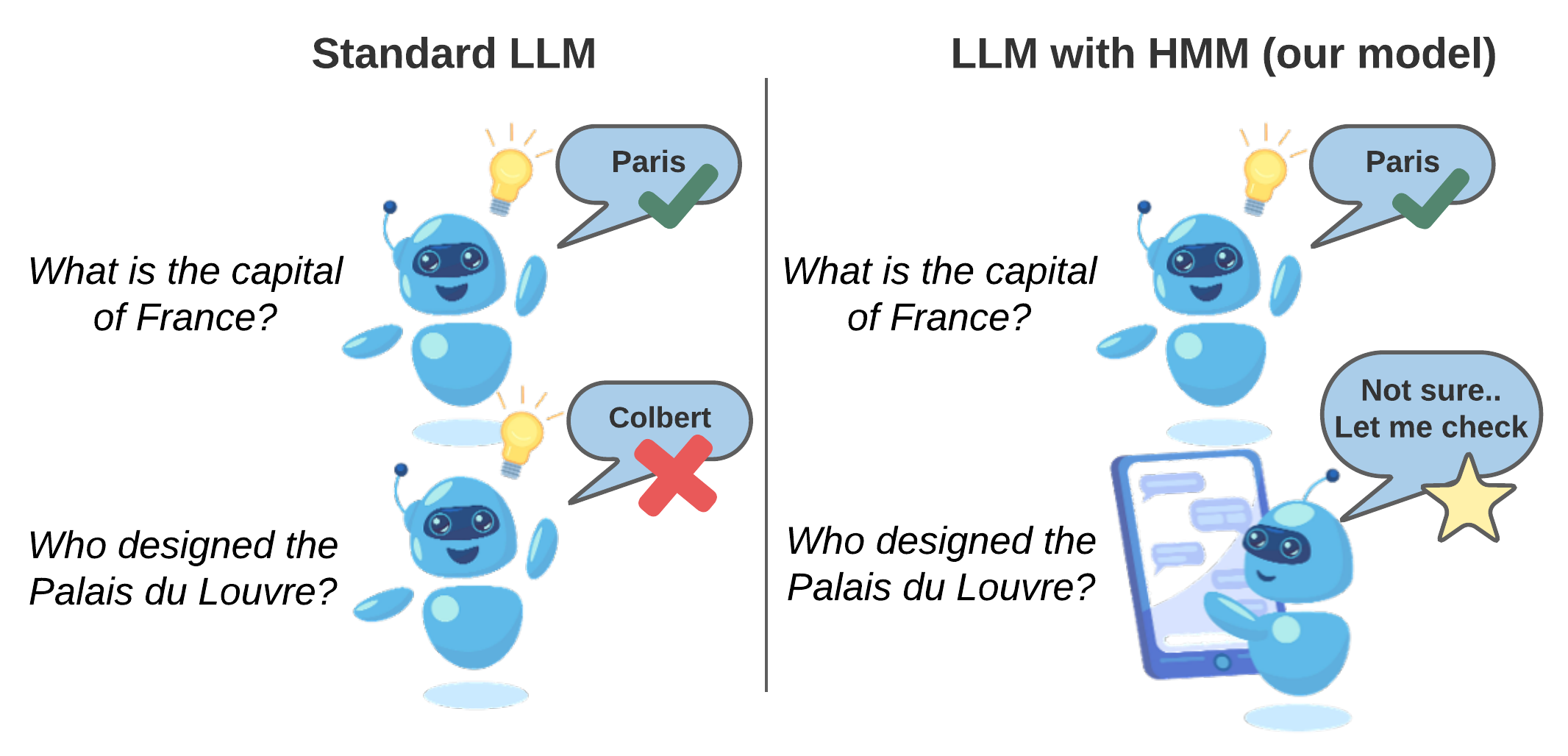}
    \caption{Our objective is to  teach an LLM to trade off between accessing an external tool and answering directly. The LLM should access an external knowledge base only when it is not self-confident about the answer, otherwise, it should generate the answer directly, minimizing the cost of accessing an external search engine.}
    \label{fig:task}
\end{figure}

Language models have demonstrated remarkable performances in a wide range of Natural Language Processing (NLP) tasks, including conversational agents, summarization, translation, and question-answering \cite{gpt3,spark,emergent}. As scaling these models increases their ability to incorporate more and more knowledge \cite{gpt3,palm},  instead of relying on traditional search engines,  Metzler et al.  \cite{rethinkingsearch} suggest using LLM as a unified knowledge base able to perform question answering as well as document retrieval. However, even the larger models \cite{gpt3} are prone to producing inaccurate or false responses, commonly known as hallucinations \cite{10.1145/3571730}. These have been extensively explored in various NLP tasks, including summarization and translation \cite{NEURIPS2021_e4d2b6e6,guerreiro-etal-2023-looking,manakul2023selfcheckgpt,10.1145/3571730,lee-etal-2021-towards}.
Numerous approaches have been suggested to tackle this problem, with all of them employing external techniques to detect and mitigate hallucinations. In question answering, retrieval augmented methods such as REALM \cite{realm}, RAG \cite{rag}, or RETRO \cite{retro,orqa}, were proposed to reduce LLM's hallucinations. These approaches consist in grounding LLM with a retriever model to add context from a large corpus of documents and to generate answers. These architectures are effective as they both improve factualness and reduce hallucinations for specific knowledge-intensive tasks such as Open-domain Question Answering \cite{orqa}. 
However, retrieved documents are always considered without consideration of their helpfulness in solving the task. In a second line of work, models, such as LaMDA,  BlenderBot, WebGPT, Toolformer     \cite{thoppilan2022lamda,nakano2022webgpt,shuster2022blenderbot,schick2023toolformer} are specifically trained to generate a query and rely on a search engine when confronted with questions. 
While these LLMs accumulated a lot of knowledge during pre-training, they are fine-tuned to rely on external databases for each question, without considering the model’s inherent ability to answer the question. \textbf{ Toolformer calls the web API for almost all the questions, $99.3\%$ with no real discernment between directly answerable questions and the real need for external knowledge.} LLMs have accumulated a lot of information and may be able to answer directly when confronted with widely known facts \cite{gpt3,spark,emergent}. In this paper, we study a more nuanced approach that leverages external knowledge while also incorporating LLMs' intrinsic knowledge. We, therefore, propose a model that either generates a natural language answer or an API call (e.g. $\langle search \rangle$) only when the model is not self-confident about the answer, minimizing the dependency on external resources helps to save inference time and computational costs.  We focus on closed-book question-answering (CBQA) tasks and carried out on two datasets (Natural Questions (NQ) \cite{kwiatkowski-etal-2019-natural} and TriviaQA (TQA) \cite{joshi-etal-2017-triviaqa}). We study how LLMs perform at self-estimating their ability to correctly answer factual questions.

\section{Learning when to search with LLMs}

\paragraph{\textbf{Problem Formalization.}}

We consider the set of factual questions $Q$ and a set of all possible answers $\Omega$. Each question $q \in Q$ is associated with a set $A$ of correct answers with $A \subset \Omega $. In CBQA, models are able to answer factual questions without supporting documents, this means relying only on the knowledge stored in parameters. Formally, we define an LLM that maps answers with questions: $LM_{\theta}: Q \mapsto \Omega$. Let's consider an LLM that generates an answer $\hat{a}$ given a question $q$, with $\hat{a}$ that might be in the set $A$ of ground truth or not (in this last case the answer is hallucinated). 
Our objective is to train an LLM to query external resources instead of generating hallucinations or to generate directly the answers, otherwise:
\begin{equation}
    LM_{\theta'}: Q \mapsto \Omega \cup\{ \langle search \rangle\}
\end{equation}
where $\langle search \rangle$ is a call to an external tool such as a search engine. 
Training this LLM of parameter $\theta'$ can be seen as a budgeted open QA task in which we minimize the probability of generating a wrong answer $P(\hat{a} \notin A| \theta',q)$ and the probability to call an external tool $P(\hat{a} = \langle search \rangle| \theta',q)$ with a budget $\lambda$:

\begin{equation}
    \underset{\theta}{\text{argmin}} \prod_{q\in Q}  \left[  P(\hat{a} = \langle search \rangle| \theta',q)  + \lambda  P(\hat{a} \notin A| \theta',q) \right]
    \label{eq:budget}
\end{equation}

where $\lambda \ge 1$ is a hyper-parameter controlling the relative importance between accessing the external resource and a hallucination behavior. The objective of this formulation is to encourage the model to provide direct answers whenever possible, therefore minimizing the cost of searching in an external resource.  In this paper, we limit this analysis to $\lambda = 1.0$ (Equation \ref{eq:budget}). A natural solution to label the model's hallucinations is to verify whether the model outputs are correct and factual. However, the ability to answer correctly to a question is inherent to the model's size and training. There is therefore no fixed dataset with supervision labels identifying when to call an API. We, therefore, propose to leverage a language model fine-tuned on a QA dataset to infer pseudo labels from language model performance during the training. More particularly, we aim to teach LLMs to generate a special sequence of tokens $(\langle search \rangle)$, instead of answering incorrectly, without deteriorating the ability of the model to answer questions thanks to a "Hallucination Masking Mechanism".

\paragraph{\textbf{Hallucination Masking Mechanism}}
\label{sec:halluMechanism}
Our objective here is to update the model $LM_{\theta}$ to display similar performances on CBQA tasks while also detecting hallucinations. 
 Given $LM_{\theta}$ a model able to perform a CQBA task, we learn parameters $\theta'$ such that $LM_{\theta'}:  Q \mapsto  \Omega \cup\{ \langle search \rangle\}$ where the LLM can still perform CBQA but predict the $\langle search \rangle$ token instead of hallucinating an answer. 
 We introduce $\psi$ a Hallucination Masking Mechanism (HalM) allowing to mask wrong answers with $\langle search \rangle$ tokens. Formally, $\psi \circ LM_{\theta}:  Q \mapsto \Omega \cup\{ \langle search \rangle\}$:
\begin{equation}
    \psi (LM_\theta(q)) = 
\begin{cases}
    \mathds{1}(\hat{a}), & \text{if } \hat{a} \in A \\
    \langle search \rangle,         & \text{otherwise}
\end{cases}
\end{equation}

With $\mathds{1}$ the identity function and $\langle search \rangle$ the sequence of tokens used to query an external knowledge base. $\psi$ enables to generate labels for data where the identity function is applied for questions answered correctly and hallucinations are masked using  $\langle search \rangle$ tokens. This mechanism is composed on top of the original $LM_\theta$ to conserve the ability of the model to answer directly when the answer is correct. To avoid additional biases in the experiment, we limit ourselves to in-domain hallucination detection, where questions used for QA fine-tuning and hallucination detection come from the same distribution. This means that we use a single dataset for both steps, avoiding distribution shifts and shared example problems.

\section{Evaluation protocol}

\subsection{Datasets}

We consider two open-domain CBQA datasets to perform our experiments.  \\
\textbf{Natural Question Open (NQ)} \cite{lee-etal-2019-latent}: an open domain question answering benchmark derived from the Natural Question dataset \cite{kwiatkowski-etal-2019-natural} which consists of questions from web queries accompanied by a list of appropriate answers, but without the original context provided in Natural Question. \\
\textbf{TriviaQA (TQA)} \cite{triviaqa}: a dataset including questions gathered from quiz league websites and also accompanied by a list of appropriate answers.

\subsection{Metrics}
\label{section:metric}
The standard approach for assessing generative CQBA model performances is based on the consideration that a generated answer is correct if and only if it constitutes an exact match or correct answer (noted \textbf{C}) with at least one element in a list of admissible answers.
This metric alone, however, is insufficient to paint a comprehensive picture of the model's behavior, and we propose to extend it based on a comparison between the ground truth and both $LM_\theta$ and $LM_{\theta'}$'s predictions. Model output can be associated with three distinct events. A model prediction is either Correct (noted \textbf{C}), incorrect (noted \textbf{H} for Hallucinated), or a query to an external tool, namely 
$\langle search \rangle$, (noted $S$). 
As aforementioned, $LM_\theta$ predictions can only correspond to \textbf{C} or \textbf{H} events, while $LM_{\theta'}$ predictions can also correspond to $S$ type events. Following these considerations, we define true positive (TP), false positive (FP), true negative (TN), and false negative (FN) events as shown in Table \ref{tab:metric}, and use them to report F1-scores in the results. 
\begin{table}[t]
\centering
\begin{tabular}{cc|cccc}
\hline
                     &   & \multicolumn{3}{c}{$LM_{\theta'}$} &  \\
                     &   & C      & H      & S     &  \\ \hline
\multirow{2}{*}{$LM_\theta$} & C & TP     & FP     & FN    &  \\
                     & H & TP     & FP     & TN    &  \\ \hline
\end{tabular}
\caption{Table showing how predictions are considered. The $LM_\theta$ is the language model after the fine-tuning on the CQBA datasets and $LM_{\theta'}$ after the second one, namely including the HalM (Section \ref{sec:halluMechanism}).
}
\label{tab:metric}
\end{table}

\subsection{Model architectures and fine-tuning}
We consider sequence-to-sequence (encoder/decoder) models \cite{seq2seq} with different sizes to assess how scale might affect the generated data, and therefore performances. All experiments utilize both the large and XXL T5-SSM models \cite{raffel2020exploring} (770M and 11B parameters, respectively) specifically trained for CBQA using Salient Span Masking (SSM) \cite{realm}. In addition, these models have official checkpoints that were fine-tuned on NQ and TQA, saving us the computational cost of training them ourselves. 
Large models are used FP32, however, 11B parameters models are quantized into int8 to fit on GPUs.

Models are finetuned on the train of each dataset to perform traditional CBQA.
The dev set is then used to perform the second HalM-based finetuning step. This ensures high CBQA performances. Specifically, we focus on detecting hallucinations within the domain of interest, using questions from the same distribution as those used for QA fine-tuning. By utilizing a single dataset for both steps, we mean to avoid issues related to distribution shifts and shared example problems.

\begin{table*}[t]
    \begin{adjustbox}{width=\linewidth,center}
\begin{tabular}{lcccccccc}
\hline
                    & \multicolumn{3}{c}{NQ}                         &               & \multicolumn{3}{c}{TQA}                               &               \\ \hline
                    & $C$ $(\uparrow)$ & $H$ $(\downarrow)$ & Search & F1            & $C$ $(\uparrow)$ & $H$ $(\downarrow)$ & Search        & F1            \\ \hline
T5-Large            & 27.3             & 72.7               & 0.0    & -             & 19.4             & 80.6               & 0.0           & -             \\
T5-Large + PPL-t    & 18.6 (68.1\%)    & 8.9 (12.2\%)       & 72.5   & \textbf{67.7} & 12.7 (65.4\%)    & 12.7 (15.7\%)      & 74.6          & \textbf{56.4} \\
T5-Large-HalM (FT)   & 15.7 (57.5\%)    & 7.1(9.7\%)         & 77.2   & 62.4          & 14.4 (74.2\%)    & 29.9 (37.0\%)      & 55.8          & 45.0          \\
T5-Large-HalM (LoRA) & 21.3 (78\%)      & 16.6(22.8\%)       & 62.0   & 65.0          & 13.6 (70.1\%)    & 27.2 (34.8\%)      & 59.2  & 45.1          \\ \hline
T5-XXL              & 35.2             & 64.8               & 0.0    & -             & 51.9             & 48.1               & 0.0           & -             \\
T5-XXL + PPL-t      & 21.7 (61.6\%)    & 12.1 (18.6\%)      & 66.3   & 65.4          & 27.7 (53.3\%)    & 24.6 (51.1\%)      & 47.7          & 63.1          \\
T5-XXL-HalM (LoRA)   & 23.4  (66.5\%)   & 15.9 (24.5\%)      & 60.7   & \textbf{66.1} & 28.0 (53.9\%)    & 24.4   (50.7\%)    & 47.6          & \textbf{63.5} \\ \hline

Mistral-7B (16 shots)   & 28.8  (91.7\%)  & 49.9 (72.5\%)   & 21.4   & 50.6 & 34.1  (51.8\%)   & 14.7 (42.9\%)    & 51.2      & 60.0 \\ 
Mistral-7B-instruct  & 2.40 & 2.95 & 94.65  & - & 25.9 & 13.7 & 60.3  &  \\ \hline
\end{tabular}
\end{adjustbox}
\caption{Table showing the distribution of Exact match, Hallucination, and search sequence for NQ and TQA dataset. ($\%$) are showing the remaining fraction of the same behavior (C, H, S) regarding predictions of the base model. }

\label{table:distribution}

\end{table*}

\begin{table*}[t]
    \centering 
    \begin{adjustbox}{width=0.8\linewidth,center}
    \begin{tabular}{c|ccccccccccc}
    
    Ratio of correct S & 0.0 & 0.1 & 0.2 & 0.3 & 0.4 & 0.5 & 0.6 & 0.7 & 0.8 & 0.9 & 1.0 \\
    \hline
    C & 21.3 & 27.5 & 33.7 & 39.9 & 46.1 & 52.3 & 58.5 & 64.7 & 70.9 & 77.1 & 83.3 \\
    H & 78.6 & 72.4 & 66.2 & 60.0 & 53.8 & 47.6 & 41.4 & 35.2 & 29.0 & 22.8 & 16.6 \\
    \hline
    \end{tabular}
\end{adjustbox}
\caption{A table illustrating the fluctuation in hallucination and correct answer rates as the accuracy of Search S varies}
\label{tab:SCH}
\end{table*}

\subsection{Baselines and model variants}

For our models based on T5-Large and T5-XXL models,  we consider two strategies to fine-tune with  HalM: 1) the standard fine-tuning   (\textbf{FT}) and 2) using Low-Rank Adaptation (\textbf{LoRA}) \cite{hu2022lora}. Due to computational constraints, the XXL (11B) T5 model is only fine-tuned with LoRA. 
 For training LoRA, we used PEFT \cite{peft} and Adapter-transformers \cite{pfeiffer2020AdapterHub} libraries to plug the parameters efficient method to LLMs and consider a warmup with a ratio of 0.1, $r=16$, $alpha=32$, and a learning rate of $1e-4$ and $7e-5$ for large and XXL models.\\

We compare our model variants to different baselines:\\
\textbf{T5-Large and T5-XXL:} the models fine-tuned on the train set of the CBQA task. Note that these models have not been trained to call external API, solely to generate answers.\\ 
\textbf{T5-Large+PPL-t and T5-XXL+PPL-t:}  which is the strongest exogenous hallucination detection method known in the literature \cite{lee-etal-2021-towards,guerreiro-etal-2023-looking}, based on a perplexity threshold. This heuristic assesses the model's output's perplexity score and classifies it as a hallucination if it exceeds a predefined, data-derived threshold. \\
\textbf{Mistral-7B\footnote{https://huggingface.co/mistralai/Mistral-7B-v0.1}:} an in-context learning strong model with  16 examples randomly extracted from the train set. Wrong answers are masked with the 'search' sequence. These are examples of how the model should behave and be used for in-context learning. We observed that if the number of masked hallucinations and direct answers is unbalanced in the prompt, this also leads to very unbalanced prediction. Thus, we used a balanced set of examples. \\
\textbf{Mistral-7B-instruct\footnote{https://huggingface.co/mistralai/Mistral-7B-Instruct-v0.1}:} a strong instruction-based LLM prompted to follow this instruction: \textit{Answer to the question only if you know the answer, otherwise answer "I don't know"} followed by the question.\\

\section{Results}
\label{sec"res}

Table \ref{table:distribution} shows the results of all model variants and baselines for Natural Question (NQ) and TriviaQA (TQA) datasets, according to the F1-score, and proportions of correct answers \textit{C}, Hallucination \textit{H} and search \textit{S} as defined in section \ref{section:metric}. Every rate on adapted models is accompanied by the remaining fractions for each behavior (C, H, S) regarding predictions of the first model.

From a general point of view, all hallucination reduction strategies (PPL-t and HalM) are able to reduce hallucinations regarding the models fined-tuned without consideration of searching on an external API. We notice that for the  T5-Large variants the PPL-t (Perplexity-threshold) strategy outperforms the HalM with a high search rate; that might be costly.  Our variant T5-Large-HalM fine-tuned with LoRA seems to have a better balance between accurate answer generation and the search rate.  
By focusing on the T5-XXL architecture, we show that LoRA consistently outperforms PPL-T on both NQ and TQA datasets for most of the metrics. 
Indeed, the LoRA strategy retains a higher fraction of correct answers on the NQ dataset. For the TQA dataset, both approaches exhibit similar behaviors, with a slight advantage for LoRA, and manage to filter out around half the Hallucinations while retaining a similar amount of Correct answers. Altogether, these results support the claim that our proposed approach enables LLMs to endogenously identify their potential for hallucination better than perplexity-based methods.

To understand the benefit of this mechanism, we consider different correct answer ratios for the search. If the ratio is set to $0.0$, all API calls return only incorrect answers, while a ratio of $1.0$ returns only correct answers. Table \ref{tab:SCH} shows the variation of this rate for the T5-Large-HalM (LoRA) on NQ. We can see that the user experience heavily relies on the boost of correct answers provided by the search. With a ratio of $0.1$, this has similar performances as the T5-Large on NQ. While a ratio of $1.0$ provides $83.3\%$ of correct answer while the model only searches for $62\%$ of the questions.



Focusing on in-context learning models, one can see that the Mistral-7b does not perform very well with an F1-score of $50.6\%$ on NQ. Additionally, the model is very sensitive to the balance of examples in the prompt.
Regarding the instructed capabilities with Mistral-7b-instruct, we observed that the model catastrophically failed to perform the given task as the model outputs $94.65\%$ 'I don't know' while only having $2.4\%$ correct answers on NQ. This suggests that abilities outlined in LLM, namely instructions and in-context learning, are not consistent with the specific behavior to identify uncertainties without devoted fine-tuning.


\section{Conclusion}

We introduced a new model to teach an LLM to internally assess its ability to answer properly a given query, without using anything more than data used for its training. The resulting model can directly identify its ability to answer a given question, with performances comparable -if not superior- to widely accepted hallucination detection baselines such as perplexity-based approaches which are strong exogenous baselines.
In addition, this approach enables large language models to condition their generation on their ability to answer appropriately on a given query, a crucially important feature in the Toolformer approach that can learn to search only when needed.  In future work, we plan to assess the impact of the $\lambda$ hyperparameter in the  Hallucinations risk/Search trade-off.

\section*{Appendix}
\subsection{Implementation Details}
We used existing  checkpoints trained for closed-book QA \cite{roberts-etal-2020-much}. These checkpoints are available on hugging face hub \footnote{https://huggingface.co/}. To infer our label, we follow \cite{roberts-etal-2020-much} and use greedy decoding. To classify if a prediction is correct in generative QA, predictions are compared using Exact Match against a list of ground truths (GT). Because the list of GT is not exhaustive, a relative amount of predictions are False Negative, introducing noise in the training data for the second fine-tuning. For example, if the model generates "Napoleon I" but the GT only contains "Napoleon", the answer is considered False. To mitigate this, the model predictions are compared with the list of normalized ground truth: all values are lowercased and  stopwords and punctuations are removed. 
We used PEFT \cite{peft} and Adapter-transformers \cite{pfeiffer2020AdapterHub} libraries to plug the parameters efficient method to LLMs. Large models are used FP32, however, 11B parameters models are quantized into int8 to fit on GPUs. For training LoRA, we used a warmup with a ratio of 0.1, $r=16$, $alpha=32$, and a learning rate of $1e-4$ and $7e-5$ for large and XXL models.
Regarding TQA, a checkpoint is available for the XXL model, and the T5 Large SSM was fine-tuned by our care using the following hyperparameters used in \cite{roberts-etal-2020-much}: constant learning rate of $1e-3$ for  $10000$ steps, dropout of $0.1$ and batch size of $128$ and gradient accumulation of $8$. However, contrary to what is reported by authors in \cite{roberts-etal-2020-much}, we encounter overfitting after a few thousand steps.


\subsection{Histogram Log-perplexity Hallucinations Rate}
\begin{figure}[h]
    \centering
    \includegraphics[scale=0.4]{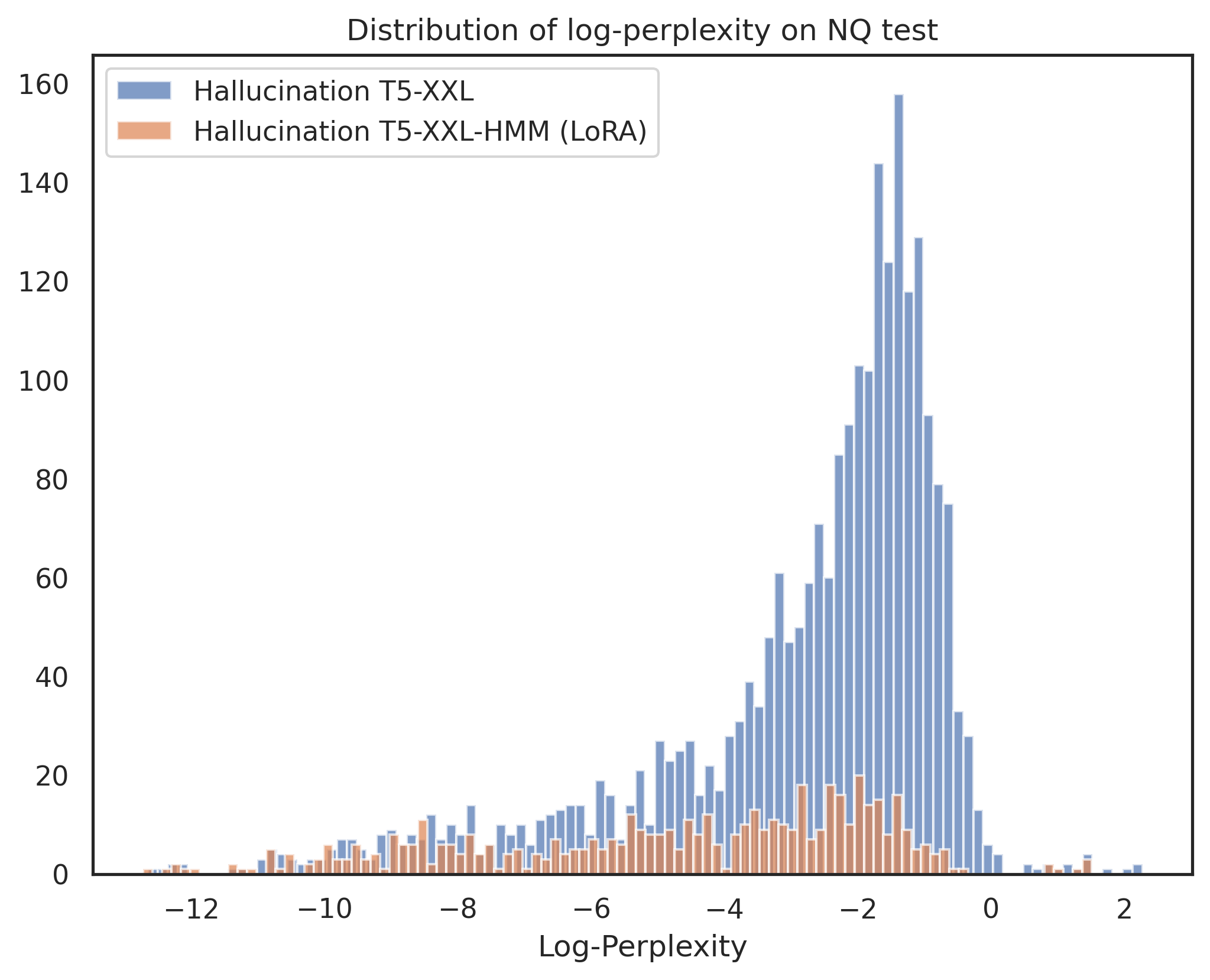}
    \caption{Distribution of hallucinations between T5-XXL and T5-XXL-HMM (LoRA) by log-perplexity.}
    \label{fig:dist}
\end{figure}

\bibliography{biblio}
\bibliographystyle{acl_natbib}

\appendix
\section{Appendix}
\label{sec:appendix}


\end{document}